# Mental Health Diagnosis in the Digital Age: Harnessing Sentiment Analysis on Social Media Platforms upon Ultra-Sparse Feature Content

*Haijian Shao, Ming Zhu, Shengjie Zhai\**

Dept. of Electrical and Computer Engineering, University of Nevada, Las Vegas, NV 89154, USA

Corresponding author: Dr. Shengjie Zhai (shengjie.zhai@unlv.edu)

*Abstract*:

Amid growing global mental health concerns, particularly among vulnerable groups, natural language processing offers a tremendous potential for early detection and intervention of people's mental disorders via analyzing their postings and discussions on social media platforms. However, ultra-sparse training data, often due to vast vocabularies and low-frequency words, hinders the analysis accuracy. Multi-labeling/Cooccurrences of symptoms may also blur the boundaries in distinguishing similar/co-related disorders. To address these issues, we propose a novel semantic feature preprocessing technique with a three-folded structure: 1) mitigating the feature sparsity with a weak classifier, 2) adaptive feature dimension with modulus loops, and 3) deep-mining and extending features among the contexts. With enhanced semantic features, we train a machine learning model to predict and classify mental disorders. We utilize the Reddit Mental Health Dataset 2022 to examine conditions such as Anxiety, Borderline Personality Disorder (BPD), and Bipolar-Disorder (BD) and present solutions to the data sparsity challenge, highlighted by 99.81% non-zero elements. After applying our preprocessing technique, the feature sparsity decreases to 85.4%. Overall, our methods, when compared to seven benchmark models, demonstrate significant performance improvements: 8.0% in accuracy, 0.069 in precision, 0.093 in recall, 0.102 in F1 score, and 0.059 in AUC. This research provides foundational insights for mental health prediction and monitoring, providing innovative solutions to navigate challenges associated with ultra-sparse data/feature and intricate multi-label classification in the domain of mental health analysis.

## 1. Introduction

Over the last decade, the prevalence and increasing impact of mental health disorders have raised a pivotal concern in global public health discourses (World Health Organization, 2021)[1]. These disorders transcend age, gender, and socio-economic statuses. However, particular populations find themselves more susceptible due to a confluence of social determinants, including but not limited to, economic deprivation, systemic discrimination, and restricted educational opportunities[2]. Such socio-economic determinants have been empirically shown to widen health disparities, particularly for marginalized groups. To elucidate, data suggests that ethnic minorities within the United States frequently encounter barriers to accessing premium mental health services, especially when compared to their non-Hispanic white counterparts[3]. Such systematic inadequacies in prevention, treatment, and access for these vulnerable demographics underscore and amplify the overarching challenges of mental health care.

Concurrently, as we navigate through the digital era, there has been a transformative ascendancy of social media platforms. These platforms have evolved beyond their foundational role as communication tools, encompassing multifarious roles ranging from news dissemination and entertainment to e-commerce and public service facilitation[4]. This paradigm shift is most evident in the massive user engagements observed on platforms like Facebook and Reddit. What is particularly salient about contemporary social media is its democratization of information, empowering users not only as passive consumers, but also as active participants and creators, fostering direct, meaningful engagements with others. In this vast ocean of user-generated content lies an untapped reservoir of potential insights for the mental health sector. Pioneering researchers have begun harnessing the capabilities of Natural Language Processing (NLP) and machine learning (ML) techniques, such as sentiment analysis, to distill actionable insights from this vast and dynamic data pool. These methodologies are poised to revolutionize early detection mechanisms for individuals grappling with mental health issues[5]. Yet, a quintessential challenge specific to social media data lies in its linguistic informality—abbreviated lexicons, syntactic anomalies, and sometimes logical inconsistencies. Such linguistic traits are especially pronounced in posts from users with mental disorders, leading to an overall reduction in the "informatic feature density". Consequently, enhancing the robustness and accuracy of sentiment analysis in such contexts demands rigorous exploration and innovation.

This manuscript embarks on this intricate journey, aiming to provide a nuanced, comprehensive understanding of sentiment analysis within the milieu of social media datasets. Through an exhaustive exploration of its applications across diverse sectors, our objectives are threefold: 1) to elucidate the latent potential of extracting vital mental health markers through an ML-driven NLP framework, 2) to set a foundational precedent for subsequent research, fostering methodological enhancements in sentiment analysis for more incisive mental health surveillance and prompt interventions, and 3) to serve as a beacon of information for the wider public, with a specific focus on vulnerable demographics necessitating augmented care. Our investigative foray is anchored in several pioneering innovations:

1) **Navigating Data Sparsity:** Leveraging a novel weak classifier, we strategically address the profound challenge precipitated by an alarming 99.81% feature space sparsity. This classifier adopts an iterative paradigm, zeroing in on and rectifying prior misclassifications, thus progressively honing classifier accuracy while enhancing feature encoding granularity and distribution.

2) **Adaptive Feature Length Processing:** Being acutely aware of the oscillations in feature sparsity and length, as well as their consequential ramifications on deeper feature abstraction for subsequent ML model training, we introduce the 'loop modulus' mechanism. This ensures methodological versatility and surgical precision.

3) **Augmented Feature Representation:** We embrace an innovative approach of feature expansion and concatenation, amplifying the depth and spectrum of feature representation. Such a methodology not only unearths latent features but also bolsters the model's resilience and adaptability across eclectic datasets.

In a bid to render our research more tangible and relatable, this paper also demystifies the intricacies of model parameters, particularly within hidden layers. By visualizing these, we offer a visceral understanding of the intricate interplay between hidden layer activations and input data, thereby facilitating judicious model structure choices that retain their efficacy across both training and practical application scenarios. Through these multifaceted contributions, we envisage this manuscript to serve as an indispensable touchstone for academics and industry experts, fostering collaborations at the confluence of sentiment analysis and mental health in our digitally interconnected era.

The manuscript's structure is meticulously curated for optimal reader engagement and understanding. Section 2 offers an in-depth appraisal of contemporary semantic analysis techniques, weighing their aptness and effectiveness across varied mental health patient demographics. Section 3 delves into our proposed ML-based NLP techniques, elucidating their potential in mental disorder detection. Section 4 delineates our dataset, juxtaposing the prediction acumen of our approach against extant ML-driven mental disorder detection methodologies. The final section, Section 5, culminates our contributions, accentuating their ramifications in the broader context of mental health surveillance.

## 2. Related Works

As aforementioned, the context of social media platforms may suffer from extreme low feature density, due to the various language irregularities, especially from those with mental disorders. To mitigate the feature sparsity, a few approaches can be exploited.

1) Data augmentation is one of the most commonly used approaches to improve ML models training outcomes. In NLP, with an augmented and more general corpora, low-frequency tokens that include critical information in the context (i.e., embedding vector) can be compensated, and thus reducing the impact of ultra-sparse encoded content. Utilizing a large-scale unlabeled dataset and a deep neural network (DNN) architecture, Zhang et al.[6] demonstrated the improvement of various language understanding tasks. Xie et al.[7] introduced unsupervised data augmentation techniques, as well as back-translation and word masking, for consistency training in NLP. Wei et al.[8] proposed easy data augmentation (EDA) techniques, such as synonym replacement, random insertion, random deletion, and synonym swapping, for text classification tasks. Sennrich et al.[9] focused on improving neural machine translation models by leveraging monolingual data and augmented the parallel training data with additional monolingual sentences to improve the performance.

However, although data augmentation can provide more samples, it may not resolve the imbalanced sample distribution in the feature space. Especially for patients suffering mental disorders, their feature token samples might disperse ultra-sparsely and/or overlap with samples from other major/normal categories.

2) Subword tokenization offers another avenue to tackle sparsity, by fragmenting

words into n-grams or root-based subwords. For instance, the term "unhappiness" can be dissected into segments like "un-", "happi-", and "ness." This approach reduces the sparsity of words, allows for richer semantic information, especially for out-of-vocabulary or low-frequency words. Sennrich et al.[10] introduced subword units, such as Byte-Pair Encoding (BPE), to reduce the sparsity and improve the translation quality when handling rare words in neural machine translation. Bojanowski et al.[11] represented words as bags of character n-grams and employed a FastText model, so as to enrich the word vector with more morphological and semantic information. Luong et al.[12] presented a multi-task learning approach for sequence-to-sequence models, which can leverage the shared subwords and alleviate sparsity issues caused by low-resource languages or rare words. Kudo et al.[13] introduced subword regularizations to adapt variations in subword segmentation, thus reducing the impact of sparsity and improving translation performance. Meanwhile, recent single process theories also explain the mirror effect that special characteristics of some low-frequency words (in terms of either letter features or semantic features) are more likely to be recognized correctly (i.e., higher hits and lower false alarm)[14]. Even though, this low-frequency advantage is affected by list composition[15,16].

On the other hand, subwords may introduce additional ambiguities and computational complexities, as subwords will significantly extend the data dimension whilst some subwords may not contain actually semantic information.

3) Reducing the dimension of word vectors, as seen in models like Word2Vec, can also palliate data sparsity. A general Word2Vec model intends to utilize high dimensional vectors to capture semantic features among words. A reduced-dimensional feature space, wherein each dimension corresponds to a distinct feature, enables enhanced semantic representation of word relationships and serves to alleviate sparsity by minimizing the number of non-zero elements and redundancy within the vector. Leveraging the tenets of the Johnson-Lindenstrauss lemm[17] and positing the existence of a linear mapping in dimensionality reduction effectively uphold all inter-point pairwise distances within the dataset, subject to a relative error. Ma et al.[18] developed an end-to-end sequence labeling model that combines bidirectional long short-term memory (LSTM), convolutional neural network (CNN), and conditional random field (CRF). By leveraging the convolutional layer, the model reduces the dimensionality of input representations to effectively address the sparsity issues. Reimers et al.[19] investigate the impact of reporting score distributions on the performance of LSTM networks for sequence tagging, and reduced dimensionality-related sparsity to improve the overall performance of the model. Jozefowicz et al.[20] incorporated bypasses in recurrent neural networks (RNNs) to alleviate dimensionality-related sparsity and enable more effective information flow. Belinkov et al.[21] discussed attention mechanisms and probing tasks that alleviate dimensionality-related sparsity by focusing on important features or subspaces within the neural models.

Nevertheless, dimension reduction may lose implicit but critical information in the original data, especially for ultra-sparse dataset (e.g., semantic dataset for mental health diagnosis). Such loss of information, unbalanced distribution, and potential noises may significantly degrade the accuracy and sensitivity in information/feature capturing for the subsequent ML model training.

   4) Alternative word vector models, such as GloVe and FastText, are developed with their distinctive representation and training strategy to combat sparsity. Pennington et al. [22] introduced the GloVe model, which combines global matrix factorization with local context window-based word co-occurrence statistics. Bojanowski et al.[11] introduced FastText to represent words as bags of character n-grams to capture morphological and semantic information of words, which is particularly useful for handling rare and out-of-vocabulary words. Feng et al.[23] introduced a pretrained BERT-like Transformer-based model (i.e., CodeBERT) that captures syntax, structure, and contextual information that are related to programming languages from a huge corpus of publicly available code repositories. Its attention layers capture long-range interdependence across the context. Mikolov et al.[24] introduced Continuous Bag-of-Words (CBOW) and Skip-gram to learn dense word vector representations upon distributional hypothesis and their contextual distributions. Peters et al.[25] introduced Embeddings from Language Models (ELMo) that generate context-dependent word representations by training bidirectional language models. ELMo addresses sparsity by capturing fine-grained syntactic and semantic information, allowing downstream models to better understand word meanings in context.

   All aforementioned methods can be used individually or in combination to address the sparsity within the word vectors. However, these word vector models typically rely on a relatively small context segment and are trained on large-scale general-use pretraining corpora or vocabulary. As a result, their performance and effectiveness of capturing semantic information, especially global contextual information, may be limited in certain specific realms, such as mental disorder diagnosis which may include a large number of rare/low-frequent words, terminologies, irregular expressions, etc. These words may lack corresponding vector representations in the general word vector model, leading to the persistence of sparsity issues and inaccuracy in understanding the context.

## 3. Proposed Advanced Classification Prediction Approach

   Initially, to facilitate nuanced data analysis and sophisticated modeling, it is imperative to subject the raw natural language data samples to a meticulous preprocessing regimen before extracting the hidden features with the subsequent ML models. This involves a series of nuanced operations to refine the textual data, such as stemming and lemmatization to reduce words to their root forms, tokenization to break down large chunks of text into individual words or phrases, and sentence segmentation to demarcate distinct statements or ideas. Further, it is crucial to eliminate linguistic noise that might detract from the analytical precision. Thus, stop words, which include special characters, punctuation marks, and common articles, are meticulously removed. In the interest of uniformity, all lexical entities are converted to lowercase. Furthermore, to uphold the integrity and privacy of the data, specific contexts, such as web URLs, alongside any personally identifiable information, such as names and phone numbers, are either obfuscated or entirely expunged. Through these meticulous preprocessing steps, the data is primed for more advanced analytical endeavors.

Then, the text content is mapped to a dimension-adjustable TF-IDF word vector[26,27] via Tfidfvectorizer, which is a commonly used vectorization technique in text mining and information retrieval. It transforms a collection of text documents into a numerical feature representation using the term frequency-inverse document frequency approach [28], which is defined as Eq. (1),

$$TFIDF = TF(t,d) \times IDF(t) \qquad (1)$$

where $TF(t,d) = \sum_{x \in d} fr(x,t)$ stands for the number of occurrences of term $t$ (i.e., term frequency, TF) in document $d$, while $IDF(t) = \log|D|/(1 + |\{d: t \in d\}|)$ represents the inverse document frequency (IDF). Noted that, $|D|$ stands for the total number of documents, $fr(x,t) = \begin{cases} 1, & if\ x = t \\ 0, & otherwise \end{cases}$, and $|\{d: t \in d\}|$ represents the number of documents that contain the term $t$. As a result, when a term exhibits a high frequency in a single document (i.e., TF) but is rare in the broader collection (i.e., IDF), it signifies its importance within that specific context. In other words, TF-IDF feature/value holds a substantial weight/implication for mental disorder identification when a single document (i.e., locally) exhibits a high TF for a specific term, coupled with a low occurrence of that term (i.e., IDF) across the entire semantic data collection (i.e., globally).

Assume that the input context is mapped into a set of word vectors with TF-IDF features: $x = [x_1, x_2, ..., x_N]$. Denote $\sigma$ as the sparsity of $x$, Eq. (2), where $\mu(x_o)$ stands for the count of 0 elements in $x$, and $\|x\|$ indicates the total number of elements in $x$.

$$\sigma = \mu(x_0)/\|x\| \qquad (2)$$

Due to the fragmented and chaotic nature of the expressions from individuals with mental disorders, their messages and content on the social platforms (e.g., Reddit) often appear irregularities in syntax, fluctuations in emotions, lack of structure and/or consistency of the context. All these factors may lead to the high sparsity (i.e., $\sigma \to 1$, over 99.8%) of critical words/informatic features within the word vector (Figure 1.a). Naturally, the higher sparsity it is, the less likelihood and/or accuracy to achieve the automated mental disorder identification. It could get worse especially when there lacks sufficient context, and/or when the patient struggles with more than one mental disorders (i.e., multi-labels). In order to effectively address the ultra-sparsity issue in mental disorder recognition in social media context, we proposed a ML-based classification approach (Figure 1) to distinguish users mental disorder status. Our approach innovatively proposes a series of techniques, namely weak classifier, loop modulus and feature enhancement, respectively, to suppress the feature sparsity that causes low data gradient, informatic feature loss, and difficulty in distinguishing among classes. Pseudocode is illustrated in Algorithm 1 and Figure 1.b, while more detailed code is provided in Supplementary. After concatenating different data features, the feature sparsity is significantly mitigated to about 85.4% (Figure 1.c). These enhanced features are then fed to train multiple ML models (Figure 1.d), and their training outcomes are evaluated upon various criteria, such as accuracy, F1 score, etc. (Figure 1.e). The evaluation results validate the efficacy of the proposed approach in classifying the mental disorders using social media content.

---

**Algorithm 1** Proposed approaches for data sparsity

1: **procedure** INPUT:$(x_{sparse}^{train,test}, y_{label}^{train,test})$ ▷ Preprocessed dataset
2: Output: $Model_{input}$
3: **for** $t$ in range($d_{length}$): ▷ Gradient classifier
4:     **if** $TF(t,d) \neq 0$:
5:         $TFIDF \leftarrow \sum_{x \in d} fr(x_{i,sparse}, t) \times \log \frac{|D|}{1+|\{d:t \in d\}|}$
6:         $x_{sparse}^{train}(t), y_{label}^{train} \leftarrow TFIDF.fit.transform\left(x_{sparse}^{train}(t), y_{label}^{train}\right)$
7:         $x_{sparse}^{train}(t) \leftarrow x_{sparse}^{train}(t).toarray()$
8: **return** $x_{sparse}^{train}(t)$
9: **for** $i, m$ in range($N; M$): ▷ Building classifiers
10:
11:     $r_{im} \leftarrow -\frac{\partial L(y_{i,label}, F_{m-1}(x_{i,sparse}))}{\partial F_{m-1}(x_{i,sparse})}$
12:
13:     $\gamma_m \leftarrow \arg\min_{\gamma} \sum_i L\left(y_{i,label}, F_{m-1}(x_{i,sparse})\right) + \gamma h_m(x_{i,sparse})$
14:
15:     $h_m(x_{i,sparse}) \leftarrow y_{i,label} - F_m(x_{i,sparse})$
16:
17:     $clf.fit\left(\{x_{i,dense}\} \leftarrow \{x_{i,sparse}\}\right)$
18:
19:     $classifier \leftarrow classifier.append(clf)$
20: **return** $classifier$
21: **for** $clf$ in $classifier$: ▷ Gradient estimation
22:
23:     $\nabla\left\{x_{dense}^{fea}\right\} \leftarrow \nabla(clf.apply(x_{dense}[:, :, 0]))$
24:
25:     $\nabla\left\{x_{dense}^{fea}\right\} \leftarrow np.hstack\left(\nabla\left\{x_{dense}^{fea}\right\}\right)$
26: **return** $\nabla\left\{x_{dense}^{fea}\right\}$
27: **for** $i$ in range($x_{dense}^{fea}.shape[0]$): ▷ Loop modulus
28:
29:     $\nabla_x x_{dense}^{fea}, len \leftarrow \nabla\left\{x_{dense}^{fea}[i]\right\}, \nabla\left\{x_{dense}^{fea}\right\}.shape[0]$
30:
31:     $\nabla_y x_{dense}^{fea} \leftarrow \nabla\left\{x_{dense}^{fea}[i + x_{dense}.shape[0]]\right\} \mod (len)$
32:
33:     $\nabla \times \left\{x_{dense}^{fea}\right\} \leftarrow np.concatenate\left(\nabla_x x_{dense}^{fea}.flatten(), \nabla_y x_{dense}^{fea}.flatten()\right)$
34:
35:     $poly_{1st}\left\{x_{dense}^{fea}\right\} \leftarrow ploy.fit.transform\left(x_{dense}^{fea}[i].reshape(-1, 1)\right)$
36:
37:     $Model_{input} \leftarrow \left\{x_{dense}^{fea}, \nabla x_{dense}^{fea}, \nabla \times x_{dense}^{fea}, poly_{1st}x_{dense}^{fea}, y_{label}\right\}$
38: **return** $Model_{input}$ ▷ Feature cascade

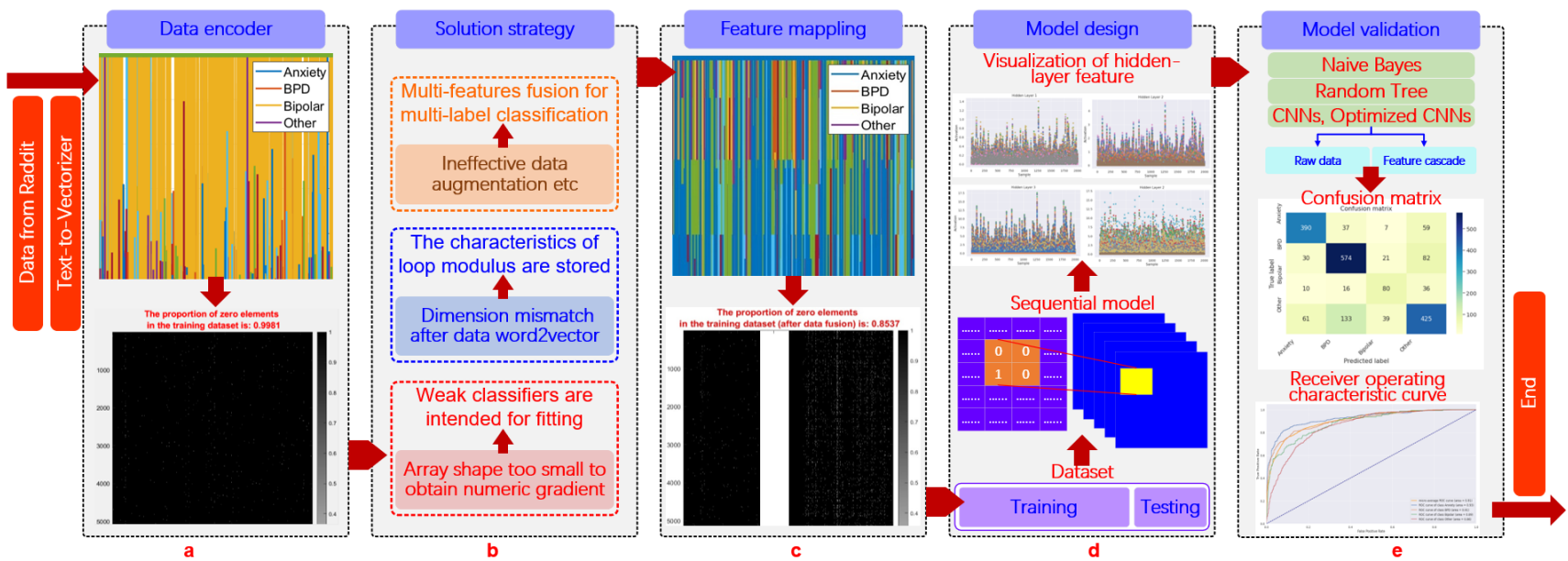

*Figure 1 The processing diagram of the proposed classification approach*

## 3.1 Weak Classifier to Improve the Feature Density

After getting the TF-IDF featured word vectors, we apply a weak classifier model $\hat{y} = F(x_{i,sparse})$ to mitigate the sparsity in the feature space. We employ a gradient boosting with a negative pseudo-residual of the loss function, Eq. (3), to put more attention on the wrongly predicted samples in each training epoch of the weak classifier.

$$r_{i,m} = -\left[\frac{\partial L\left(y_{i,label}, F_{m-1}(x_{i,sparse})\right)}{\partial F_{m-1}(x_{i,sparse})}\right]_{i=1,\ldots,N, m=1,\ldots,M} \quad (3)$$

We also set a learning rate of $\gamma_m$ as Eq. (4), so that it is adaptive to the gradient and the residue of the loss function.

$$\gamma_m = \arg\min_{\gamma} \sum_i L(y_{i,label}, F_{m-1}(x_{i,sparse}) + \gamma \cdot r_{i,m}) \quad (4)$$

After each training epoch $m$, the errors of the predecessor model (i.e., $L\left(y_{i,label}, F_{m-1}(x_{i,sparse})\right)$ are expected to be reduced. Assume that after $m \to M$ epochs, the training loss converges whilst the validation process also reaches its optimum, we achieve the enhanced feature set with lower sparsity, i.e., $\{x_{i,sparse}\}_{i \to N} \xrightarrow{m \to M} \{x_{i,dense}^{fea}\}$.

## 3.2 Loop Modulus for Feature Analysis

Due to the uneven sparsity of the raw data, extracted feature lengths may vary significantly. As such, we propose to segment the extracted feature using loop modulus. Assume $\{x_{i,dense}^{fea}\} \subset \mathbb{R}^{u \times v}$ stands for the enhanced features of each sample, with $u$ for the number of sample classes and $v$ for the maximum feature lengths, respectively. We set up a dual-loop modulus algorithm for feature re-organization. The outer loop iterates over the number of samples in the *X_train* data array $\{x_{dense}^{fea}\}$ (i.e., *X_train.shape*[0]). The inner loop first calculates the gradient value $\nabla_x x_{dense}^{fea} = \nabla\{x_{dense}^{fea}[i]\}$ of each sample. Secondly, we calculate the gradients array for $\nabla_y x_{dense}^{fea}$ based on the index $(i + X_{train}.shape[0])$. For index that exceeds the range of $x_{dense}^{fea}.shape[0]$, we apply a modulo operation %($train\_gradients.shape[0]$) to wrap around, Eq. (5). This index modulation not only resolves the uneven distribution of feature lengths due to the data/feature sparsity, but also provides a self-adaptive solution to the changing feature dimensions with extended generalizability to the entire feature extraction approach.

$$idx_{mod} = (i + X\_train.shape[0]) \ \% \ train\_gradients.shape[0] \quad (5)$$

## 3.3 Feature Enhancement for ML-Based Mental Disorder Classifications

In this study, we calculate another three derivative operators, namely gradient $\nabla(\cdot)$, curl $\nabla \times (\cdot)$ and first-order polynomial $poly_{1st}(\cdot)$, upon the densified feature spaces $\{x_{i,dense}^{fea}\}$ for the subsequent ML model training. Generally speaking, gradient represents the intensity of value changes over the space, while curl implies the local context and possibly interconnection relationship among words and sentences in NLP. Curl can be calculated by concatenating the aforementioned $\{\nabla_x x_{dense}^{fea}\}$ and $\{\nabla_y x_{dense}^{fea}\}$. Meanwhile, $poly_{1st}(\cdot)$ generates another feature dimension as a complement. Statistically, these derivatives of these functional operators (i.e., $\nabla(\cdot)$, $\nabla \times (\cdot)$, $poly_{1st}(\cdot)$) share low correlations, and thus can mutually compensate and support each other for the subsequent ML training process. Therefore, all these derivatives are flattened and concatenated as in Eq. (6), and would be employed to train ML models for mental disorder classifications.

$$Model_{input} = \{\{x_{dense}^{fea}\}, \nabla(x_{dense}^{fea}), \nabla \times (x_{dense}^{fea}), poly_{1st}(x_{dense}^{fea}), y_{lable}\} \quad (6)$$

By concatenating both linear and non-linear features, the training data is enriched with multiple uncorrelated features, so that the trained ML model can present higher accuracy as well as robustness to more generalized data, i.e., lower overfitting risks.

## 4. Experiments and Evaluations

## 4.1 Dataset Description

In this work, we employ the Mental Disorders Identification on Reddit[29] published in 2022, covering relatively general perspectives of users' real-time feelings and mental status, as well as their regions, demographics, religions, educational background, and so on. The dataset includes a total of 700,000 rows of text retrieved from multiple subreddits related to common mental disorders with 4 labels/classes: Borderline Personality Disorder (BPD), Bipolar Disorders, Anxiety Disorder, and "Others". The entire dataset was split to 0.8:0.2 for training vs validation process, respectively. Both training and validation datasets approximately share the same sample ratios for each class.

## 4.2 Feature Extractions

As discussed in Section 3, after the data pre-processing, we first establish one Gradient Boosting Classifier for each mental status class (i.e., BPD, BD, AD, and others), respectively. Then, we train each individual classifier with raw dataset and the corresponding labels, respectively. Afterwards, upon the trained classifiers, their gradient information will be stacked and used to map the ultra-sparse raw data into a relatively denser feature set. In the end, the derivatives of the dense feature set (i.e., Eq. (6)) are calculated and stacked for the subsequent ML model training. Two correlation heatmaps (Figure 2) demonstrate the low correlations among the dense feature set and its derivatives. As such, these features present little redundancy and thus are expected to compensate for more comprehensive model training.

### 4.3 ML Model Training, Evaluations and Discussions

In order to evaluate our approach in dealing with ultra-sparse semantic analysis, we have trained and tested a variety of ML models on two scenarios: 1) using original unprocessed message data (i.e., raw data, RD), 2) employing the weak classifier to improve the feature density, and the feature extraction and fusion to enhance the feature quality (i.e., feature enhanced, FE). We implemented 4 different ML models, namely Native Bayes, Random Forest (RF), a CNN, and an optimized Attention Neural Network, to compare the classification performances upon the dataset and extracted features.

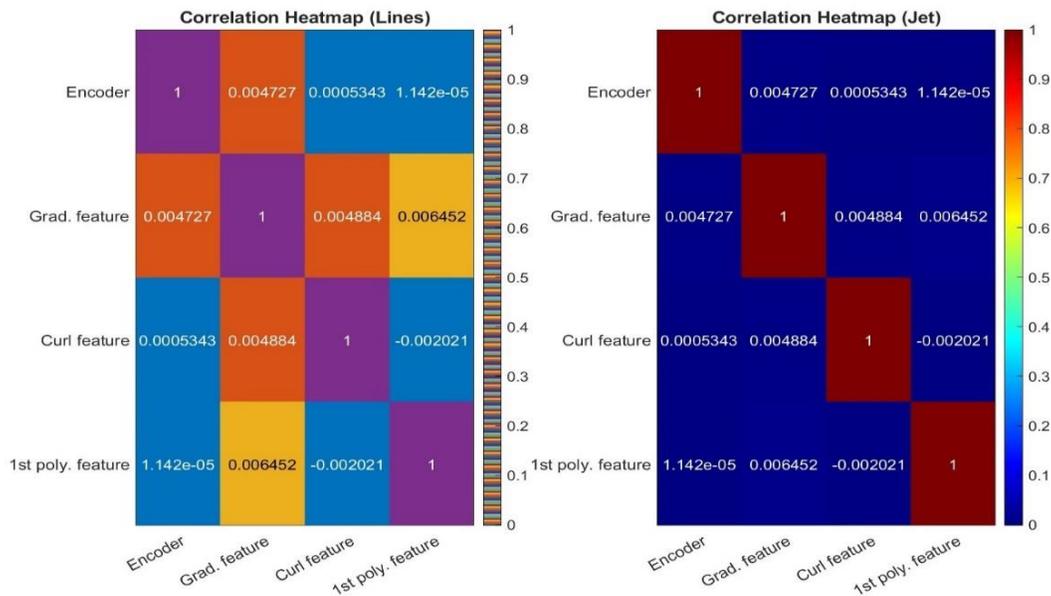

*Figure 2 The correlation heatmaps among the dense feature set and its derivatives.*

Table 1 lists the performance evaluation results for 5 different ML models for two data scenarios. One can see that RF, CNN and attention model achieves comparable and relatively optimal results. This is because these models are able to reveal the implicit features/co-relations in the context and are more adaptive to potential unbalanced data by nature. During our experiments in optimizing CCN, it turns out that compared to using deeper and more sophisticated CNNs, a relatively simple CNN with only 2 fully-connected layers (FCN) performs better in reducing the overfitting in training data, and thus improve the overall prediction/classification performance for a more generalized data. One major reason is due to the ultra-high sparsity of the features. The performance of the attention model is also similar to the optimized CNN. This could also partially be due to the lack of regularity and sometimes logic in the syntax and context, which impedes the attention module to further capture the correlation among the language based on the incorrect training results. The detailed confusion matrix and corresponding AUC-ROC of each mental disorder classification model are presented in Supplementary Figure 5-Figure 12.

Table 1. Evaluation of ML models on multi-label/class mental disorder predictions

| Model | Acc. (%) | Class | Pre. | Rec. | $F_1S$ | AUC |
|---|---|---|---|---|---|---|
| Native Bayes (RD) | 58.8% | Anxiety | 0.712 | 0.600 | 0.651 | 0.880 |
| | | BPD | 0.977 | 0.368 | 0.534 | 0.860 |
| | | bipolar | 0.505 | 0.366 | 0.424 | 0.810 |
| | | others | 0.467 | 0.862 | 0.605 | 0.740 |
| | | average | 0.665 | 0.549 | 0.554 | 0.820 |
| Random Forest (RD) | 67.8% | Anxiety | 0.758 | 0.730 | 0.744 | 0.900 |
| | | BPD | 0.694 | 0.738 | 0.716 | 0.860 |
| | | bipolar | 0.800 | 0.113 | 0.198 | 0.840 |
| | | others | 0.608 | 0.696 | 0.649 | 0.820 |
| | | average | 0.715 | 0.569 | 0.577 | 0.860 |
| CNN (RD) | 65.9% | Anxiety | 0.733 | 0.706 | 0.719 | 0.860 |
| | | BPD | 0.687 | 0.723 | 0.704 | 0.840 |
| | | bipolar | 0.400 | 0.296 | 0.340 | 0.790 |
| | | others | 0.618 | 0.635 | 0.627 | 0.800 |
| | | average | 0.609 | 0.590 | 0.598 | 0.850 |
| Attention (RD) | 66.1% | Anxiety | 0.729 | 0.671 | 0.699 | 0.860 |
| | | BPD | 0.707 | 0.694 | 0.701 | 0.840 |
| | | bipolar | 0.594 | 0.268 | 0.369 | 0.800 |
| | | others | 0.588 | 0.704 | 0.640 | 0.790 |
| | | average | 0.654 | 0.584 | 0.602 | 0.850 |
| Native Bayes (FE) | 65.8% | Anxiety | 0.671 | 0.688 | 0.679 | 0.880 |
| | | BPD | 0.728 | 0.645 | 0.684 | 0.810 |
| | | bipolar | 0.792 | 0.134 | 0.229 | 0.790 |
| | | others | 0.593 | 0.761 | 0.667 | 0.830 |
| | | average | 0.696 | 0.557 | 0.565 | 0.830 |
| Random Forest (FE) | 71.8% | Anxiety | 0.782 | 0.755 | 0.768 | 0.920 |
| | | BPD | 0.785 | 0.728 | 0.756 | 0.890 |
| | | bipolar | 0.628 | 0.415 | 0.500 | 0.870 |
| | | others | 0.632 | 0.743 | 0.683 | 0.850 |
| | | average | 0.707 | 0.660 | 0.677 | 0.880 |
| CNN (FE) | 73.9% | Anxiety | 0.744 | 0.824 | 0.782 | 0.930 |
| | | BPD | 0.890 | 0.683 | 0.773 | 0.900 |
| | | bipolar | 0.731 | 0.401 | 0.518 | 0.880 |
| | | others | 0.639 | 0.809 | 0.714 | 0.870 |
| | | average | 0.751 | 0.679 | 0.697 | 0.910 |
| Attention (FE) | 74.1% | Anxiety | 0.759 | 0.805 | 0.781 | 0.940 |
| | | BPD | 0.884 | 0.690 | 0.775 | 0.920 |
| | | bipolar | 0.652 | 0.423 | 0.513 | 0.890 |
| | | others | 0.643 | 0.815 | 0.719 | 0.870 |
| | | average | 0.735 | 0.683 | 0.697 | 0.920 |

Table 2. Average performance improvement of models using FE vs. RD

| Model | Acc. (%) | Pre. | Rec. | F$_1$S | AUC |
|---|---|---|---|---|---|
| Native Bayes | 7% | 0.031 | 0.008 | 0.011 | 0.01 |
| Random Forest | 4% | -0.008 | 0.091 | 0.100 | 0.02 |
| CNN | 8% | 0.142 | 0.089 | 0.099 | 0.06 |
| Attention | 8% | 0.081 | 0.099 | 0.095 | 0.07 |

On the other hand, for individual each model, training ML models with the proposed enhanced features can significantly improve the training outcomes in a variety of evaluation criteria (Table 2). The accuracy can be improved by up to 8%, while F1-score by about 10% and area under the receiver operating curve (AUC-ROC) by about 0.07. Specially, for both individual mental disorder classes and the average values for all 4 classes, higher F1-scores and AUC indicate the lower false positive and/or negative rate, presenting the higher confidence of the model's prediction performance, as well as the robustness of the model in correspondence to a more generalized, and potentially imbalanced dataset. Although there appear a small portion of decreases in some of these criteria when applying feature enhancement, it might be due to the ambiguities and confusion/chaotic in the raw data, as well as the imbalance and potential bias in the dataset, leading to the misunderstanding/errors in feature capture/enhancement. Nevertheless, most samples are more accurately predicted and classified, and the overall performance is improved for all models.

In addition, we visualize the trainable parameters in the hidden layers of both CNN (Figure 3) and the attention model (Figure 4). One can see that, if trained with raw data, there exhibits obvious and excessive repetitions and redundancies among each layer, preventing the effectiveness and accuracy in models' feature capturing to distinguish among different classes. On the other hand, after pre-processing the raw data with the proposed feature enhancement, the hidden layers demonstrate obvious discretion within the parameters/neurons of each layer, as well as those among different layers for different classes. Thus, we confirm that feature enhancement can significantly improve the mental disorder prediction/classification in using forum discussions.

## 5 Discussion and Conclusions

In this study, we have delved into the significant challenge of discerning manifestations of mental disorders as subtly woven into user-generated semantic content on social media platforms. The language employed on these platforms frequently features unconventional and irregular usage of words, logical constructs, and expressions. This phenomenon is particularly pronounced among individuals grappling with mental health issues, where their communications exhibit a juxtaposition of high redundancy in expression alongside a starkly diminished semantic feature density, in contrast to more formally constructed informational documents. Consequently, these ultra-sparse features present formidable hurdles in the application of NLP to analyze social media content with the aim of understanding and detecting mental health disorders.

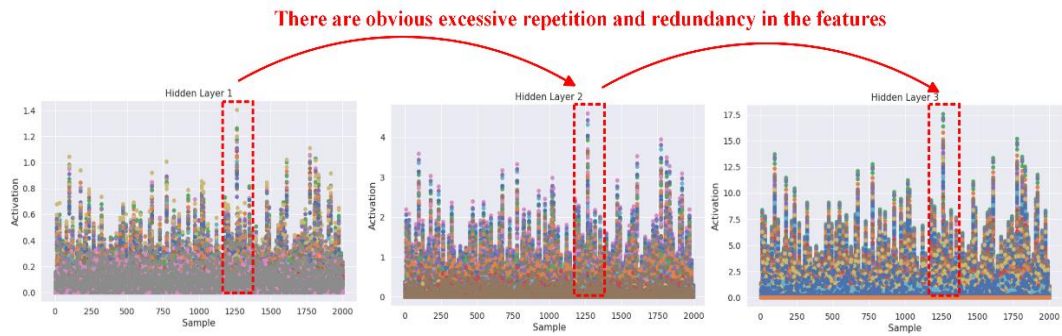

Figure 3. Visualization of hidden-layer of the CNN trained by RD (top) and FE (bottom)

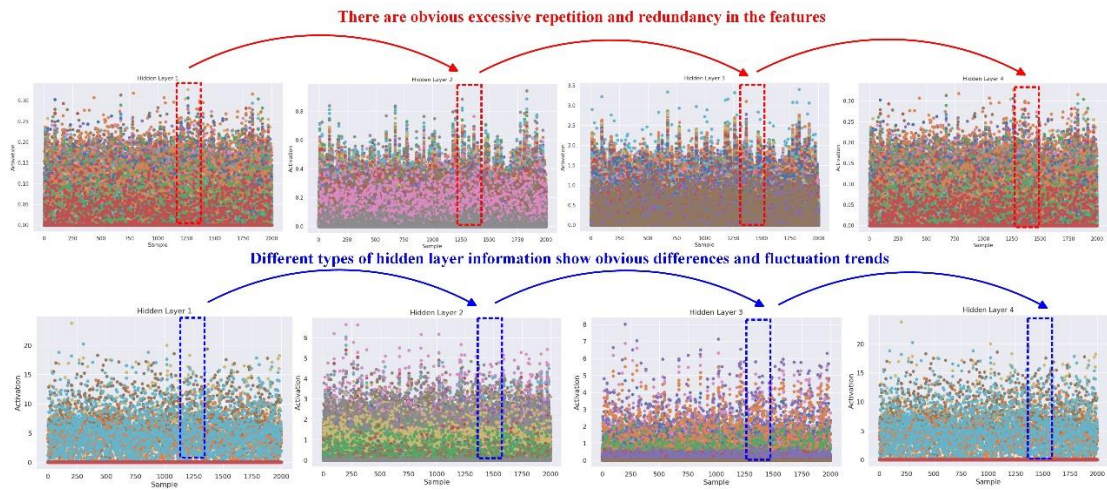

Figure 4. Visualization of hidden-layer of the attention model trained by RD (top) and FE (bottom)

To surmount the impediments posed by the paucity of dense features, we have proposed an array of pioneering methodologies. These include the implementation of weak classifiers, the incorporation of loop modulus, and a strategic enhancement of features, all tailored to bolster both the density and quality of features. This, in turn, significantly enhances the proficiency of ML models in identifying and categorizing mental disorders through textual data gleaned from social media. Our research leverages the Reddit Mental Health Dataset 2022 to appraise the efficacy of our approach in predicting four distinct mental disorders. Our findings underscore the superiority of our proposed methodology in detecting and differentiating among the specified mental disorders. We observed an average enhancement of 8.0%, 0.069, 0.093, 0.102, and 0.059 in terms of accuracy, precision, recall, F1-score, and AUC-ROC, respectively, when compared against existing NLP techniques and ML models. Furthermore, our approach facilitates a robust visualization of neural network parameters, which substantiates our approach's capability and resilience in discerning various mental disorders.

This study illuminates the promising intersection of advanced NLP techniques and mental health surveillance in the realm of social media, laying a robust foundation for future research endeavors and practical applications in this emerging discipline.

# Supplementary

*Detailed code for Algorithm 1*

---

**Algorithm 1** Proposed approaches for data sparsity
---
1: **procedure** INPUT: (**Xtrain, Xtest, ytrain, ytest**) ▷ Reddit dataset after preprocessing
2: Output: **fea-train, fea-test**
3: # ***** **Weak classifier intended for fitting** *****
4:   tfidf = $TfidfVectorizer(maxfeature = 2500, mindf = 2)$ ▷ Words appeared in at least 2 documents used as features in 2500-dimensional TF-IDF vector.
5:   **Xtrain, Xtest** = tfidf.$fit\_transform$ (**Xtrain, Xtest**) .$toarray()$ ▷ Convert a sparse matrix into a dense matrix.
6:   **NumClaTr, NumClaTe**, classifiers = **ytrain, ytest**.shape[1], [ ] ▷ Defining categories and initializing classifiers.
7: # ***** **Train a gradient classifier for class prediction** *****
8:   **for** $i, j$ **in range(NumClaTr, NumClaTe):**
9:     $clf$ = GradientBoostingClassifier()
10:    $clf$.fit(**Xtrain, ytrain**[:, $i, j$])
11:    classifiers.append($clf$)
12:    **return** classifiers
13: # ***** **Calculate the first-order gradient of the data** *****
14:   TrainGradients, TestGradients = [ ], [ ] ▷ Initialization
15:   **for** $clf$ **in classifiers:**
16:    TrainGradients.append(clf.apply(**Xtrain**)[:, :, 0])
17:    TestGradients.append(clf.apply(**Xtest**)[:, :, 0])
18:    TrainGradients = np.hstack(TrainGradients) ▷ Horizontally stack arrays or sequences together along their columns
19:    TestGradients = np.hstack(TestGradients) ▷ Horizontally stack arrays or sequences together along their columns
20:    **return** TrainGradients, TestGradients
21: # ***** **Loop modulus used for feature sorted** *****
22:   gradient_arr, vorticity_arr, poly_arr = [ ], [ ], [ ] ▷ Initialize arrays
23:   **for** $i$ **in range(Xtrain.shape[0]):**
24:    gradient_x = TrainGradients[$i$] ▷ Gradients
25:    specified_len = TrainGradients.shape[0]
26:    gradient_y = TrainGradients[($i$ + **Xtrain**.shape[0]) $mod$ specified_len] ▷ Use modulo to wrap around
27:    grad_con = np.concatenate((gradient_x.flatten(), gradient_y.flatten()))
28:    gradient_arr.append(grad_con)
29:    vorticity = compute_vorticity(gradient_x, gradient_y) ▷ Vorticity
30:    vorticity_arr.append(vorticity.flatten())
31:    poly = PolynomialFeatures(degree=1) ▷ ploy feature
32:    polys = poly.fit_transform(**Xtrain**[$i$].reshape(1, -1))
33:    poly_arr.append(polys)
34:   gradient_arr = np.array(gradient_arr)
35:   vorticity_arr = np.array(vorticity_arr)
36:   poly_arr = np.array(poly_arr)
37:    **return** gradient_arr, vorticity_arr, poly_arr
38:         ▷ Repeat the preceding steps for **Xtest** to be **gradient_arr** etc.
39: # ***** **Data fusion based on constructed features** *****
40:   **fea_train** = np.concatenate((**Xtrain**, gradient_arr, vorticity_arr, poly_arr), axis=1) ▷ Repeat the preceding steps for **Xtest** to be **fea_test**.

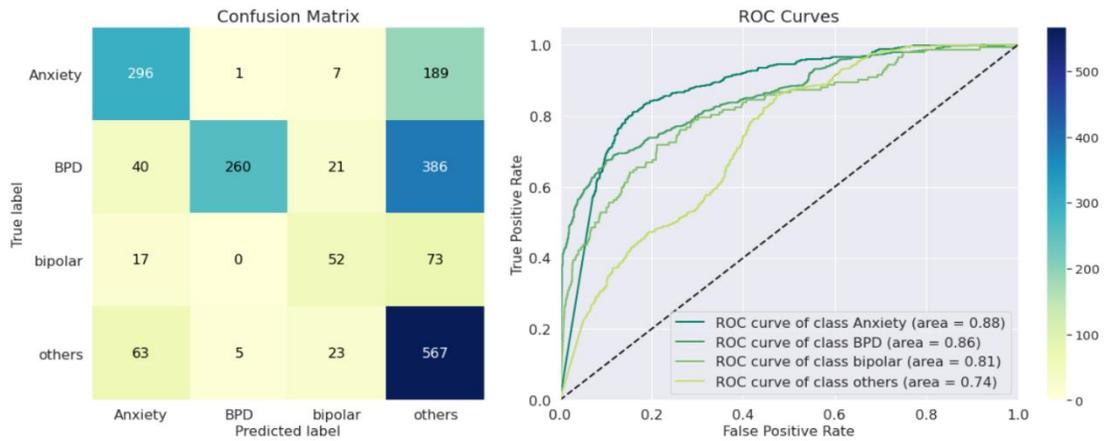

Figure 5. Confusion matrix and ROC Curve of Native Bayes (Raw data)

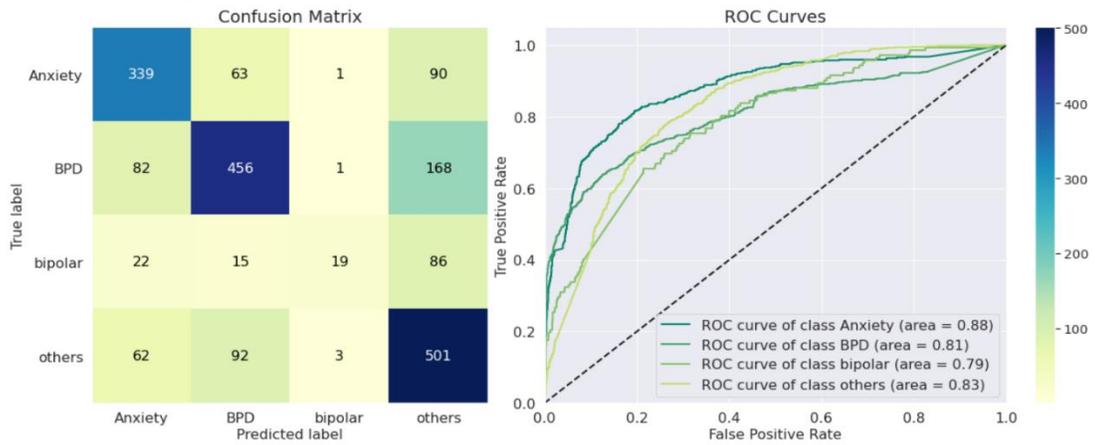

Figure 6. Confusion matrix and ROC Curve of Native Bayes (Feature enhanced)

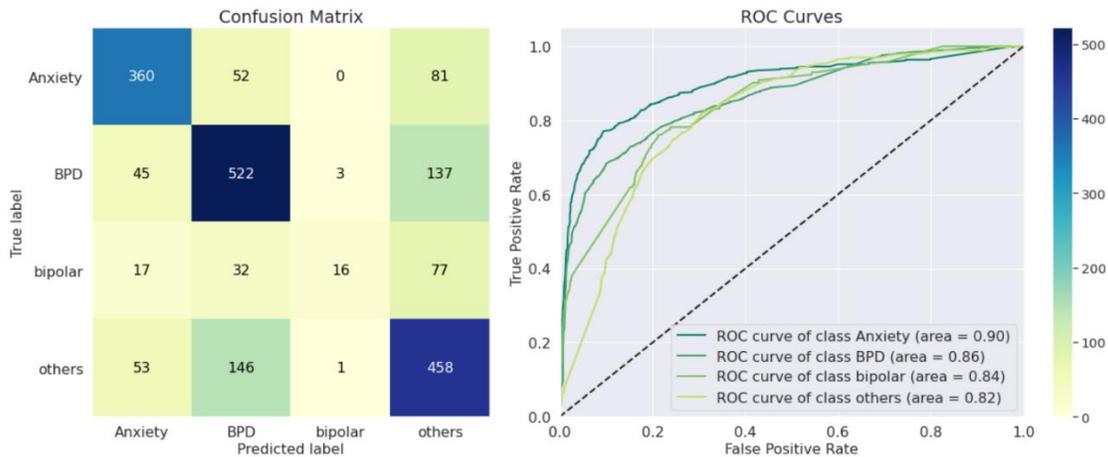

Figure 7. Confusion matrix and ROC Curve of Random Forest (Raw data)

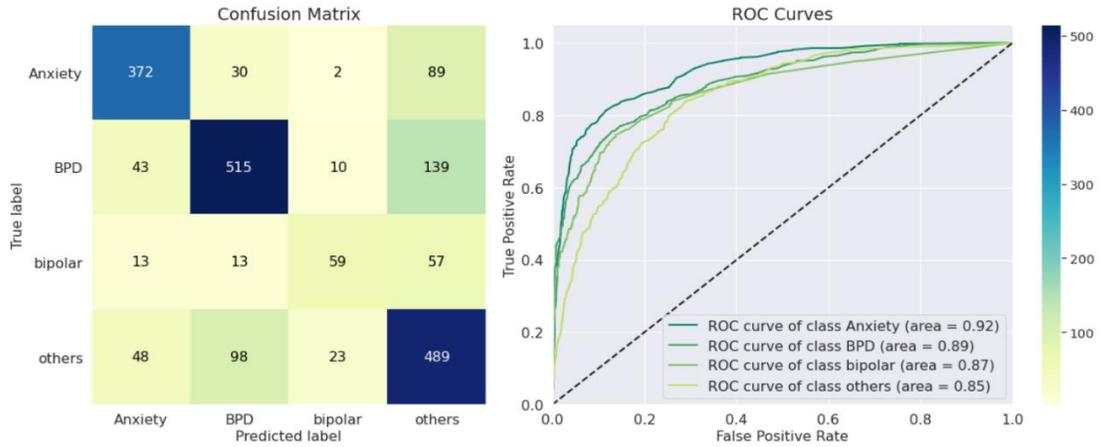

Figure 8. Confusion matrix and ROC Curve of Random Forest (Feature enhanced)

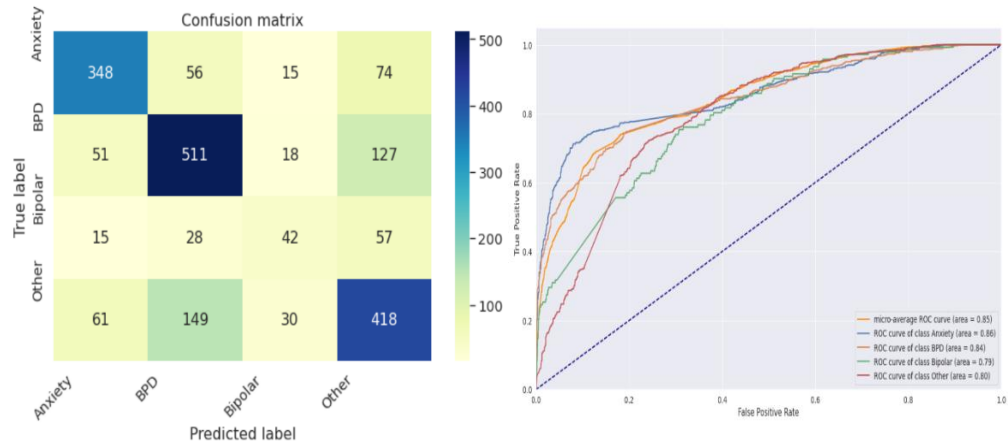

Figure 9. Confusion matrix and ROC Curve of the CNN model (Raw data)

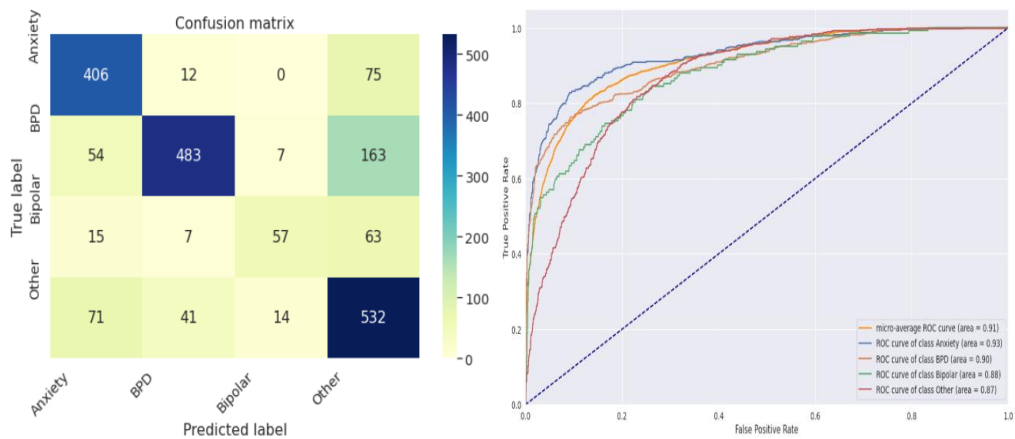

Figure 10. Confusion matrix and ROC Curve of the CNN model (Feature enhanced)

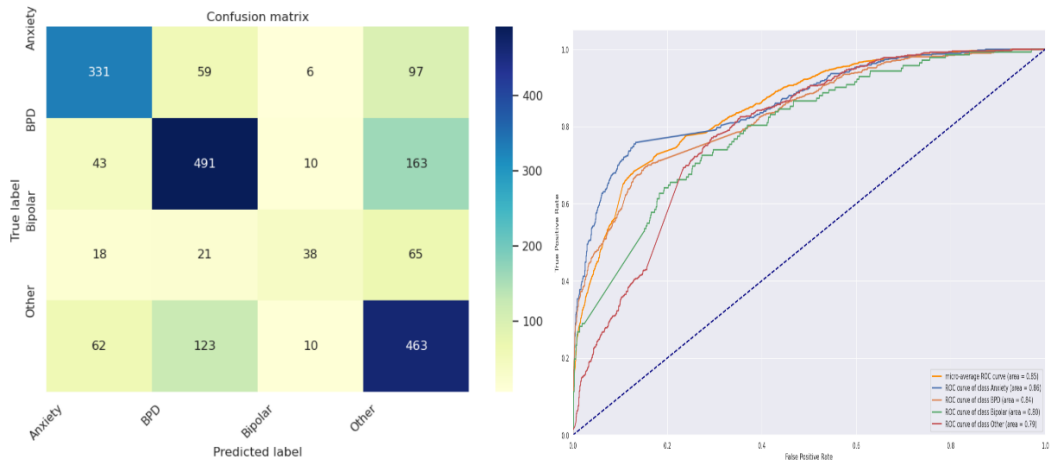

Figure 11. Confusion matrix and ROC Curve of the attentional model (Raw data)

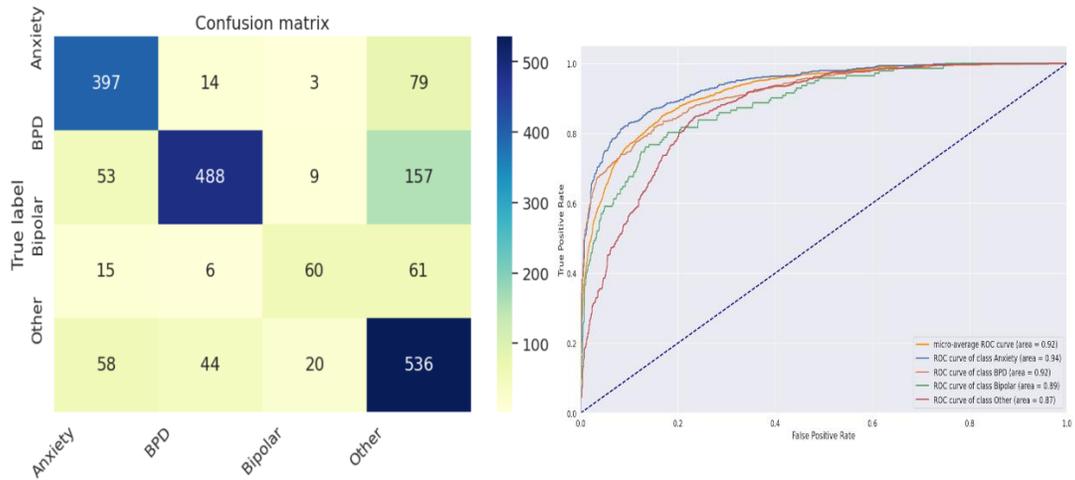

Figure 12. Confusion matrix and ROC Curve of the attentional model (Feature cascade)